\documentclass[letterpaper, 10 pt, conference]{ieeeconf}  

\IEEEoverridecommandlockouts                              

\usepackage{cite}
\usepackage{amsmath,amssymb,amsfonts}
\usepackage{wasysym}
\usepackage{comment}
\usepackage{graphicx}
\usepackage{textcomp}
\usepackage[dvipsnames]{xcolor}
\usepackage{multirow}
\usepackage[export]{adjustbox}
\usepackage[nolist,nohyperlinks]{acronym}
\usepackage{url}
\usepackage{booktabs}
\renewcommand{\baselinestretch}{0.965} 

\DeclareUnicodeCharacter{2212}{\textendash}

\title{\LARGE \bf
Optimizing Interaction Space: Enlarging the Capture Volume for Multiple Portable Motion Capture Devices
}

\author{Muhammad Hilman Fatoni$^{1,2**}$, Christopher Herneth$^{1**}$, Junnan Li$^{1}$, Fajar Budiman$^{1,2}$,\\
Amartya Ganguly$^{1}$, and Sami Haddadin$^{1}$ 
\thanks{** equal contribution}%
\thanks{$^{1}$The authors are with the Munich Institute of Robotics and Machine Intelligence (MIRMI), Technische Universität München (TUM), Germany. $^{2}$The authors are also with Institut Teknologi Sepuluh Nopember (ITS), Indonesia.
        {\tt\footnotesize {\{muhammadhilman.fatoni, christopher.herneth, junnan.li, fajar.budiman, amartya.ganguly, and haddadin\}@tum.de}}}
}

\begin{document}

\maketitle

\thispagestyle{empty}
\pagestyle{empty}


\begin{abstract}

Markerless motion capture devices such as the \ac{LMC} have been extensively used for tracking hand, wrist, and forearm positions as an alternative to \ac{MMC}. However, previous studies have highlighted the subpar performance of \ac{LMC} in reliably recording hand kinematics. In this study, we employ four \ac{LMC} devices to optimize their collective tracking volume, aiming to enhance the accuracy and precision of hand kinematics. Through Monte Carlo simulation, we determine an optimized layout for the four \ac{LMC} devices and subsequently conduct reliability and validity experiments encompassing 1560 trials across ten subjects. The combined tracking volume is validated against an \ac{MMC} system, particularly for kinematic movements involving wrist, index, and thumb flexion. Utilizing calculation resources in one computer, our result of the optimized configuration has a better visibility rate with a value of 0.05 $\pm$ 0.55 compared to the initial configuration with -0.07 $\pm$ 0.40. Multiple \ac{LMCs} have proven to increase the interaction space of capture volume but are still unable to give agreeable measurements from dynamic movement.

\end{abstract}




\begin{acronym}

\acro{ADLDAT}[ADL Dataset]{ADL Human Arm Motion Dataset}
\acro{IQR}[IQR]{Interqartile range}
\acro{FRoM}[FRoM]{Functional Range of Motion}
\acro{RoM}[RoM]{Range of Motion}
\acro{CoM}[CoM]{Center of Mass}
\acro{dof}[DoF]{Degree of Freedom}
\acro{adl}[ADL]{activities of daily living}
\acro{dulm}[MoBL-ARMS DULm]{MoBL-ARMS Dynamic Upper Limb model}
\acro{tpm}[TPm]{Transhumeral Prosthesis model}
\acro{LMC}[LMC]{Leap Motion Controller}
\acro{LMCs}[LMCs]{Leap Motion Controllers}
\acro{MMC}[MMC]{Marker-based Motion Capture}
\acro{FoV}[FoV]{Field of View}
\acro{SVD}{Singular Value Decomposition}
\acro{PSO}{Particle Swarm Optimization}
\acro{SDK}{Software Development Kit}
\acro{UDP}{Universal Datagram Protocol}
\acro{NaN}{Not a Number}
\acro{MCP}{Metacarpophalangeal}
\acro{DIP}{Distal Interphalangeal}
\acro{PIP}{Proximal Interphalangeal }

\end{acronym}

\section{Introduction} 
\label{sec: Introduction}

Marker-based motion capture (MMC) serves as the gold standard for recording human kinematics \cite{Sheiko2022}. In clinical settings, it proves invaluable for assessing movement accuracy, identifying injury risk factors \cite{Hogue2023}, and personalizing diagnostic approaches for conditions like stroke, Parkinson's disease, cerebral palsy, spinal cord injury, and multiple sclerosis \cite{Vincent2022}.

However, \ac{MMC} systems are expensive, time-consuming to set up and calibrate, and not commonly available while requiring trained personnel \cite{Cotton2023markerless_Cocap}. In clinical settings, where patient comfort and convenience are paramount, portable and low-cost markerless motion capture provides a more sensitive tool for research and rehabilitation \cite{Meng_Bíró_Sárosi_2023}. Leap Motion controller\textsuperscript{TM} (LMC) represents one such markerless motion capture device that utilizes infrared cameras and a proprietary internal model for tracking hand and finger movements. Despite its low cost and ease of use, a single \ac{LMC} device did not demonstrate acceptable agreement with marker-based motion capture \cite{ganguly2021comparison, ganguly2021can}.

\begin{figure}[tb]
    \centering
    \fontsize{6.5pt}{6.5pt}\selectfont
    \def\svgwidth{0.5\textwidth}
    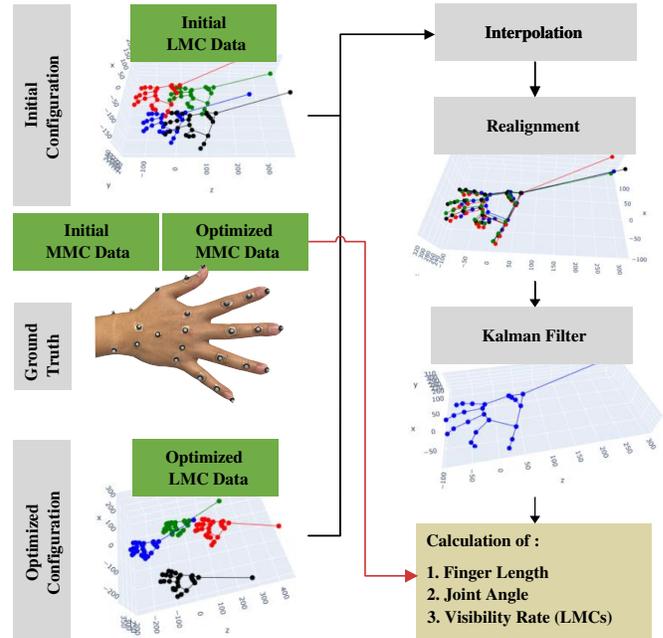
    \vspace{-0.6\baselineskip}
    \caption{Multiple \ac{LMCs} validation pipeline for initial and optimized \ac{LMC} placement. \ac{LMC} data is captured - interpolated - realigned - Kalman filtered. Comparisons are made to ground truth optical marker trajectories from \ac{MMC}.}
    \label{fig: Method}
    \vspace{-2.2\baselineskip}
\end{figure}

To overcome this limitation, Novacek et al. \cite{NovacekProjectMultileap} fused angle estimations of multiple \ac{LMC}s. The Kabsch algorithm \cite{Kabsch1976} was used with weights assigned to recordings of individual LMCs based on confidence levels computed from palm orientation angles. However, their method only gauged confidence levels for the entire hand, while credence in individual fingers was not considered. In this study, a ray-tracing algorithm was proposed, estimating the visibility of individual finger phalanges. Occlusions of virtual markers in \ac{LMC} measurement frames were computed based on simplified geometric hand models and estimated hand configurations. Resulting finger occlusion metrics and measurements of hand properties from multiple \ac{LMCs}' marker fused by a Kalman filter \cite{KalmanFilter}.

\begin{figure*}
    \centering
    \includegraphics[width=0.85\linewidth]{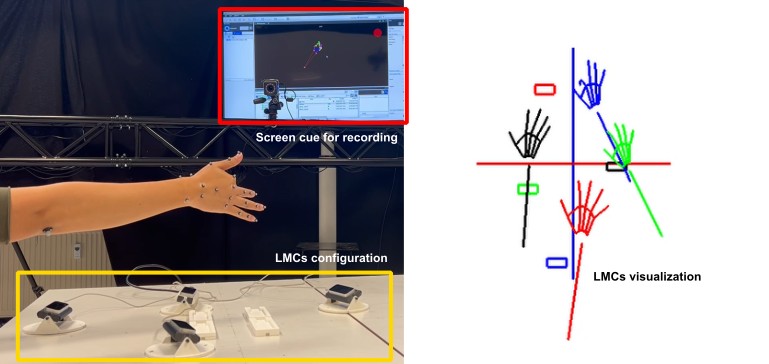}   
    \vspace{-0.6\baselineskip}
    \caption{Experiment setup. Left: Simultaneous recording of \ac{MMC} and \ac{LMCs} system. Markers were placed on the hand while an experiment of index flexion movement in the vertical pose of optimized \ac{LMCs} configuration was running. The cue recording is a big red circle shape, informing the subject to start the trial. Right: Visualization of \ac{LMCs} marker reading using OpenGL in developed custom program.}
    \label{fig: Experiment Setup}
    \vspace{-1.5\baselineskip}
\end{figure*}

However, challenges persist, particularly regarding the reliability of capture volume and tracking performance inherent to the \ac{LMC}. Consequently, optimized LMC positions were computed for multiple \ac{LMCs}, based on Monte Carlo Simulation \cite{MonteCarloSimulation} of hand trajectories expected in trials. \ac{LMC}s were located and oriented such that the visibility of as many fingers by as many \ac{LMC} as possible was ensured in each trial frame.  

Figure \ref{fig: Method} shows the complete pipeline of this study. Our contributions are as follows: (1) an online multi-\ac{LMC} framework, fusing measurements of multiple devices where occlusions metrics calculated in the ray-tracing algorithm, (2) an \ac{LMC} placement optimization, computing ideal \ac{LMC} placement of multiple devices based on \ac{LMC} frame samples expected during trials, (3) validation against gold standard \ac{MMC}, demonstrating the effectiveness of our approach. In summary, this study alleviates the limitations of singular \ac{LMC} devices and posits a solution for markerless motion capture relevant for rehabilitation and clinical measurements to facilitate reliable diagnosis for hand impairments. 
\section{Materials and Methods} 
\label{sec: Methods}

\subsection{Subject Information}
Ten right-handed male subjects with a mean age of 28.90$\pm$4.95 years participated in this study. The subjects gave their written informed consent. The study is registered with ethics serial number 2022-588-S-KH. It was conducted according to the guidelines of the Declaration of Helsinki and approved by the Ethics Committee of the Technical University of Munich. 

\subsection{Markerless Motion Capture System}
The \ac{LMC} is a markerless portable motion capture device (Ultraleap, Bristol, UK). \ac{LMC} consists of two 640x240-pixel near-infrared cameras separated 40 mm apart and three LEDs spaced on either side and between the cameras. The dimensions of \ac{LMC} are 13 mm in height, 80 mm in width, and 30 mm in depth, weighing around 32 grams. Although \ac{LMC} is extensively used in gaming, several studies have utilised \ac{LMC} and robots simultaneously \cite{Weichert2013, Du2015AMarkerless}. \ac{LMC}  typically operates at 120 Hz \cite{Ultraleap:UH-003206-TC}, but the actual sampling operation is a variable sampling \cite{ganguly2021comparison, Smeragliuolo2016Validation, Niechwiej-Szwedo2018-Evaluation}. \ac{LMC} has an interaction zone of about 60 cm, extending from the device in a 120x150$^{\circ}$ field of view. It can track 28 marker positions of the hand from the elbow to the tip of the fingers. Previous research \cite{houston2021evaluation, ganguly2021comparison} used these marker positions to measure hand anthropometrics.

Recent developments of \ac{LMC} have made it possible to read multiple \ac{LMCs} simultaneously. The marker data read from \ac{LMC} following a structured format. They consisted of point coordinates, which are the elbow, wrist, palm, and bones of the fingers. The \ac{LMC} employed a right-handed Cartesian coordinate system \cite{ultraleapLeapConcepts}. The local origin of \ac{LMC} was located at the centre top. The x- and z-axes lie in the plane of the camera sensors, with the x-axis running along the camera baseline. The y-axis is vertical, and the z-axis points towards the user. This study utilised four \ac{LMCs} in two configurations. The first configuration was decided by placing all four \ac{LMCs} on a flat surface where all \ac{LMCs}' centre positions are placed in each corner of a box shape. This initial arrangement was placed assuming it would increase the capture volume. The second configuration was decided from the result of the optimization algorithm, which is shown in Figure \ref{fig: MultiLMC Config} (blue \ac{LMC} configuration). During the pilot testing phase of this study, \ac{LMC} hand-tracking capability was tested to determine whether the performance was within the datasheet specification. Captured marker data served as indicators to determine whether hand position within \ac{LMC} interaction space. However, it was found that at a distance of 250 mm from the origin, the value from this axis distance was cut. Because of this, an offset was introduced in the realignment step. Therefore, a second transformation matrix for realignment is deployed to correct this difference.

\subsection{Marker Based Motion Capture System}
Sixteen Vero v2.2 cameras (Vicon Motion Systems Ltd, Oxford, UK) were used for validation in this study. The system has a resolution of 2.2 MP and a maximum frame rate of 330 Hz, with a sampling frequency of 100 Hz. This frame rate was set to match the final sampling rate of four \ac{LMCs} used in this study. Twenty-six markers were carefully placed onto bony landmarks on the arm and hand of the subject. This placement is similar to \cite{ganguly2021comparison}. However, this study also placed additional markers on the elbow and wrist. The purpose of this placement is to mimic the marker position of the \ac{LMC}. On the elbow joint, the marker used was 9 mm in diameter, while the other placements were 4 mm. The placement of the marker for the recording of the \ac{MMC} as ground truth is shown in Figure \ref{fig: Method}.

\subsection{Experimental Protocol}
The subject sat comfortably on a chair behind a desk. On the top of the desk, four \ac{LMCs} were configured in two settings: initial configuration or optimized configuration. The subject put their right hand on top of the \ac{LMCs} (around 15 cm) and produced one static trial and three dynamic trials in two hand poses for each configuration. Each trial was performed five times. The motions were flexion of the index finger, the thumb, and the wrist. The subject performed two open-hand poses, one with the palm facing downward (horizontal) and the other with the palm facing to the left (vertical). These poses were chosen to assess the configuration efficacy in ensuring the visibility of at least one \ac{LMC} on the finger. Data was recorded simultaneously for both \ac{LMC} and the \ac{MMC}, as shown in Figure \ref{fig: Experiment Setup}.

\subsection{LMC Placement Optimization}
\label{sec: LMC placement optimization}

\begin{figure}[tb]
    \centering
    \fontsize{9pt}{9pt}\selectfont
    \def\svgwidth{0.4\textwidth}
    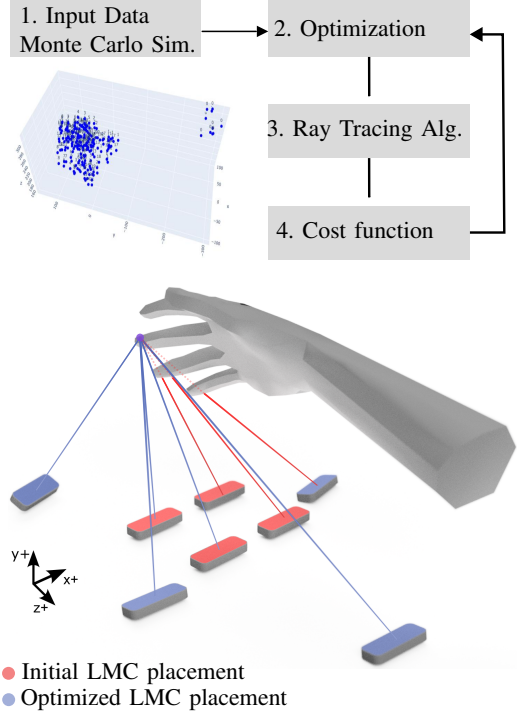
    \vspace{-0.6\baselineskip}
    \caption{Optimization procedure: 1: generation of input data; 2 - 4: optimization with ray-tracing algorithm. The bottom shows exemplary \ac{LMCs} ray-tracing for the tip of the middle finger from optimized positions (blue) and initial positions (red).}
    \label{fig: MultiLMC Config}
    \vspace{-1.5\baselineskip}
\end{figure}

The study comprises two steps. The first step involves determining the optimized configuration for four \ac{LMCs}, shown in Figure \ref{fig: MultiLMC Config}. The second step, shown in Figure \ref{fig: Method}, entails validating the optimization results through experimentation. During the first step, ten hand poses serve as reference points to generate virtual markers. These reference hand poses encompass static configurations during index, thumb, and wrist flexion in horizontal and vertical palm positions. Hand marker datasets required for optimization were generated using Monte Carlo simulation \cite{MonteCarloSimulation}, which simulated various wrist and finger configurations along expected trial trajectories. Virtual markers, representing the \ac{LMC} marker-set, were extracted from each configuration. Ray-tracing against these virtual markers of multiple \ac{LMCs} frames was used to optimize the experimental location and orientation of the four \ac{LMCs} setup. An optimal \ac{LMCs} placement was defined by the metric $\mathfrak{M}$ described in \eqref{equ: PSO metric}, where $F_i$ corresponds to the minimum of \ac{LMC}s seeing a particular finger across all fingers in marker frame $i$. At the same time, $N$ represents the number of frames in the dataset. The metric ensures prioritization of frames with invisible fingers ($F_i = 0$) while providing a transparent gradient towards increasing the number of \ac{LMCs} with unobstructed views on finger markers. This metric is used as a cost function mentioned in the optimization process in Figure \ref{fig: MultiLMC Config}.
\begin{equation}
    \mathfrak{M} = 
    \sum_{i=1}^{\#\ac{LMC}s}\sum_{j=1}^{N} \frac{F_j}{N^{i}}
    \label{equ: PSO metric}
\end{equation}
\par The visibility of individual fingers for a particular \ac{LMC} frame was determined by testing the following scenarios on the virtual marker set $M$ of the frame:
\begin{enumerate}
    \item Does the vector pointing from the subject elbow to the wrist $\boldsymbol{\vec{f_a}}$ form an acute angle with the \ac{LMC} forward direction? 
    \item Are all \ac{LMC} virtual markers with location vectors $\boldsymbol{\vec{m}}$ within the \ac{LMC} \ac{FoV} \\ 
    $\forall m \in M, k \in {1, 2, 3 ,4}: |\boldsymbol{\vec{m}}| <= L \ \wedge  \ \boldsymbol{\vec{m}} \cdot \boldsymbol{\vec{n_k}} > 0$ Here the \ac{FoV} was outlined as the set of normal vectors $\{\boldsymbol{\vec{n_0}},\boldsymbol{\vec{n_1}}, \boldsymbol{\vec{n_2}}, \boldsymbol{\vec{n_3}}\}$ describing the planes forming the inverted \ac{LMC} visibility pyramid as well as the maximal distance to the \ac{LMC} camera centre $L$.
    \item Does a ray pointing from the \ac{LMC} camera centre to any virtual finger marker intersect the palm of the same hand?
    \item Does the same ray intersect another finger before reaching the finger marker tested for visibility?
\end{enumerate}
Conditions 3 and 4 were tested for each frame and each marker belonging to a given finger. Occlusions of a single finger marker labelled the entire finger as invisible. Condition 3 was assessed by computing the regression plane \eqref{equ: regression plane} formed by the palm marker set $\{m_{P0}, m_{P1}, ..., m_{P5}\}$ and calculating the intersection point $p_1$ of the marker ($m$) - \ac{LMC} ray $\boldsymbol{\vec{r}}$ with that plane \eqref{equ: intersection point}. An occlusion occurred when the intersection point lived within the convex hull formed by the palm markers ($ \exists \ x \ s.t. \ |m_{P0}, m_{P1}, ..., m_{P5}| \cdot x \leq p_1 \ \& \ x_i > 0$), was in the \ac{LMC} \ac{FoV}: ($\arccos(p1 \cdot m) = 0$) and closer to the \ac{LMC} than the marker ($|p_1| < |m|$). 
\begin{equation}
    U \Sigma [\boldsymbol{\vec{v_1}}, \boldsymbol{\vec{v_2}}, \boldsymbol{\vec{v_3}}]^{T} = SVD(\boldsymbol{P_i} - \overline{\boldsymbol{P_i}})
    \label{equ: regression plane}
\end{equation}
\begin{equation}
    p_1 = m - \boldsymbol{\vec{r}} \cdot \frac{\boldsymbol{\vec{v_3}} \cdot (p - m_{P0})}{\boldsymbol{\vec{v_3}} \cdot \boldsymbol{\vec{r}}}
    \label{equ: intersection point}
\end{equation}
Condition 4 was formalized by constructing cylinders of 1 cm diameter along the three phalanges of each finger. A marker was invisible if the marker-\ac{LMC} ray intersected any phalange cylinder of another finger, and that intersection point was located between the \ac{LMC} and the marker. Say $p_{h1}$ and $p_{h2}$ are markers at a finger's PIP and DIP joints, and $\boldsymbol{\vec{p_h}}$ is the unit vector pointing from $p_{h1}$ to $p_{h2}$. Then the marker $m$ - \ac{LMC} ray $\boldsymbol{\vec{r}}$ intersects the phalange if the shortest distance $d$ \eqref{equ: shortest distance} between $\boldsymbol{\vec{p_h}}$ and $\boldsymbol{\vec{r}}$ was smaller than the cylinder radius ($d < 0.5$ cm), the point of shortest distance $p_2$ \eqref{equ: point of shortest distance} on $\boldsymbol{\vec{p_h}}$ lay between $p_{h1}$ and $p_{h2}$ ($|p_2 - p_{h1}| < |\boldsymbol{\vec{ph}}|$)and $p_2$ was closer to the \ac{LMC} than $m$ ($|p_2| < |m|$). 
\begin{equation}
    d = \frac{abs((\boldsymbol{\vec{p_h}} \times \boldsymbol{\vec{r}}) \cdot (p_{h1} - m))}{|\boldsymbol{\vec{p_h}} \times \boldsymbol{\vec{r}}|}
    \label{equ: shortest distance}
\end{equation}
\begin{equation}
    p_2 = p_{h1} + \boldsymbol{\vec{p_h}} \cdot \frac{(\boldsymbol{\vec{r}} \times (\boldsymbol{\vec{p_h}} \times \boldsymbol{\vec{r}})) \cdot (m - p_{h1})}{((\boldsymbol{\vec{p_h}} \times \boldsymbol{\vec{r}}) \cdot (\boldsymbol{\vec{p_h}} \times \boldsymbol{\vec{r}}))}
    \label{equ: point of shortest distance}
\end{equation}
\par Metric \eqref{equ: PSO metric} was minimized by \ac{PSO} \cite{PSO}, optimizing the placement and orientation of a total of four \ac{LMCs}. \ac{LMCs} coordinates $\{x, y, \phi, \theta\}$ represent their 2D location as well as their rotation along the \ac{LMC} long and vertical axes. 

\subsection{Data Collection}
The proposed framework employed a single computer to collect data from four \ac{LMCs}, in contrast with other studies that utilize multiple computers for this purpose \cite{WangEnlarging, Kiselev2019HandGesture, Hu2017HumanStochastic}. The computer specifications included an 11th Gen Intel(R) Core(TM) i9 processor with 32GB RAM. Preliminary investigations revealed that each \ac{LMC} requires a specific USB bandwidth, allowing for connectivity of two \ac{LMCs} per USB port. Consequently, in this study, the \ac{LMCs} were connected to the computer through two USB ports, with two \ac{LMCs} sharing the same port via a USB extender. Moreover, external power was provided to the extender to ensure sufficient current supply of all \ac{LMCs}.

A custom program\footnote{\url{https://github.com/hilmanfatoni/Multi-LMC_Optimization}} was developed to acquire and process data from these four \ac{LMCs} on a single computer. The program utilizes \ac{SDK} 5.6.1 provided by Ultraleap, capable of simultaneously reading data from multiple \ac{LMCs}. Developed in C++, it facilitates the use of the C-based \ac{SDK}. Each \ac{LMC} was identified by its serial number and named LMC1, LMC2, LMC3, and LMC4. Timestamped data from each \ac{LMC} was interpolated to ensure synchronization at 100 Hz using B-spline interpolation, with the initial timestamp defined when the \ac{LMC} first detects the hand. As noted in prior studies \cite{ganguly2021comparison, houston2021evaluation, WangEnlarging}, the \ac{LMC} system lacks a fixed frame rate, employing variable sampling instead. Standardizing this sampling was the initial data processing step. Ganguly et al. \cite{ganguly2021comparison} resampled data at 150 Hz to align with \ac{MMC}, while Houston resampled LMC data at 100 Hz \cite{houston2021evaluation}. Wang et al. also employed interpolation with a sampling interval of 0.033s \cite{WangEnlarging}. Our study found the original sampling rate, when all four \ac{LMCs} were read, to be approximately 11 - 34 Hz, lower than the specified maximum of 120 Hz. Data visualization was employed using OpenGL, and \ac{UDP} communication synchronized \ac{LMC} and \ac{MMC} system recording times. Each trial commenced when all \ac{LMCs} began recording, triggering the \ac{MMC} system recording protocol simultaneously. 

The experiment for each subject comprised 16 trial types conducted across two different \ac{LMCs} configurations and two distinct hand poses, resulting in a total of 80 trials per subject. Trial specifics are outlined in Table \ref{tab: number_experiment}. Each trial involved the recording of two datasets: \ac{LMC} data and \ac{MMC} data. All subjects completed all experiment trials except for Subject 7, who did not participate in wrist movement trials. The total dataset comprised 1560 records from both the \ac{LMC} and the \ac{MMC} System.

\begin{table}[t]
\centering
\caption{List of motion type and number of trials in experiment}
\vspace{-0.6\baselineskip}
\label{tab: number_experiment}
\begin{tabular}{@{}lc@{}}
\toprule
\multicolumn{1}{c}{\textbf{LMC Configuration}} & \textbf{Number of Trial}      \\ \midrule
\textbf{Initial Configuration}         & \multicolumn{1}{l}{}          \\
\textbf{Pose: Open hand poses horizontal}      & \multicolumn{1}{l}{}          \\
1. Static                                      & 5 times                       \\
2. Index flexion                               & 5 times                       \\
3. Thumb flexion                               & 5 times                       \\
4. Wrist flexion                               & 5 times                       \\
\textbf{Pose: Open hand poses vertical}        &                               \\
5. Static                                      & 5 times                       \\
6. Index flexion                               & 5 times                       \\
7. Thumb flexion                               & 5 times                       \\
8. Wrist flexion                               & 5 times                       \\
\textbf{Optimized Configuration}               &                               \\
\textbf{Pose: Open hand poses horizontal}      & \multicolumn{1}{l}{\textbf{}} \\
9. Static                                      & 5 times                       \\
10. Index flexion                              & 5 times                       \\
11. Thumb flexion                              & 5 times                       \\
12. Wrist flexion                              & 5 times                       \\
\textbf{Pose: Open hand poses vertical}        & \multicolumn{1}{l}{\textbf{}} \\
13. Static                                     & 5 times                       \\
14. Index flexion                              & 5 times                       \\
15. Thumb flexion                              & 5 times                       \\
16. Wrist flexion                              & 5 times                       \\ \bottomrule
\end{tabular}
\vspace{-1.8\baselineskip}
\end{table}

\subsection{Data Processing}
As shown in Figure \ref{fig: Method}, four types of marker data were recorded: Initial \ac{LMC} Data, Initial \ac{MMC} Data, Optimized \ac{LMC} Data, and Optimized \ac{MMC} Data. \ac{LMC} data went through the stages of interpolation, realignment, and Kalman filtering. For \ac{MMC} data, these processes were not carried out; instead, the process of calculating hand properties such as finger length and joint angle was calculated based on marker data. \ac{MMC}'s marker data was preprocessed by verifying marker data from the MMC system to prevent potential mislabeling or missing markers. During preprocessing, It was found that the simultaneous recording process between \ac{MMC} and \ac{LMCs} would only be problematic if the marker were too close to the \ac{LMCs}. Markers that are close to the \ac{LMCs}, for example, when performing a wrist flexion movement, will experience mislabeling because the \ac{LMC} light is recognized as a marker. For \ac{LMCs} data, they went through an interpolation process after being acquired. This process resulted in all \ac{LMCs} data sampled exactly every 10 ms and recorded in the file where markers from LMC1, LMC2, LMC3, and LMC4 were listed sequentially.

\ac{LMCs} data then underwent a realignment process. Different from other studies \cite{houston2021evaluation, NovacekProjectMultileap} where using the Kabsch Algorithm to reorient other sensors to the same reference sensor, this study uses a transformation matrix that transforms all markers relative to each \ac{LMC} to the world coordinate which located in the centre of experiment. The Kabsch Algorithm only works when both the reference sensor and the sensor to be transformed can "see" the same corresponding marker. However, in this study, such conditions are not consistently met, as there are instances where one or more \ac{LMCs} fail to detect the hand. To address this, an algorithm to handle undetected hand or a \ac{NaN} value was carried out. After realignment, all four markers from \ac{LMC} were fused using Kalman Filter \cite{KalmanFilter}. This study uses the Kalman Model from Houston \cite{houston2021evaluation}. This study assumes that markers read by \ac{LMC} follow Gaussian distribution and are linearly related to the state. A different part from the Kalman Model was modified to handle \ac{NaN} value. In the update step of Kalman Filter, \ac{NaN} marker value that comes from one of the \ac{LMCs} would be skipped and not used as part of updating the state.

Fused \ac{LMC} data and \ac{MMC} data were then calculated together to measure finger length and joint angle. The purpose of the experiment is to validate the optimized configuration against the gold standard \ac{MMC} system. The metrics used were finger length and joint angle. Additionally, the origin \ac{LMC} data underwent calculation of visibility rate. The finger length was calculated by measuring the distance of markers placed between finger segments from static trial data. The markers placement of \ac{MMC} system was configured in the exact location of \ac{LMC} markers. By doing so, the calculation result between the \ac{MMC} system and \ac{LMC} (both in initial and optimized configurations) should be the same. The joint angle was calculated by using four markers. From these four markers, every two markers will make a line. This line represents the finger segment. Then, there will be two lines formed by these four markers, and one of their ends meets. By utilizing the cosine function from those two lines, the angle of the joint was calculated. This angle calculation is adapted from \cite{ganguly2021comparison}. In this study, the joint angle was measured from the dynamic trial that focused on the movement of the index finger, thumb, and wrist. Based on that movement, the resulting joint angles are Index \ac{MCP}, Index \ac{PIP}, Index \ac{DIP}, Thumb \ac{MCP}, Thumb \ac{DIP}, and Wrist Joint. The last metric used is the visibility rate. Each \ac{LMC} from each trial would have visibility values per frame, which was calculated using a ray-tracing algorithm.
\section{Results} 
\label{sec: Results}

\begin{figure*}
    \centering
    \includegraphics[width=0.81\linewidth]{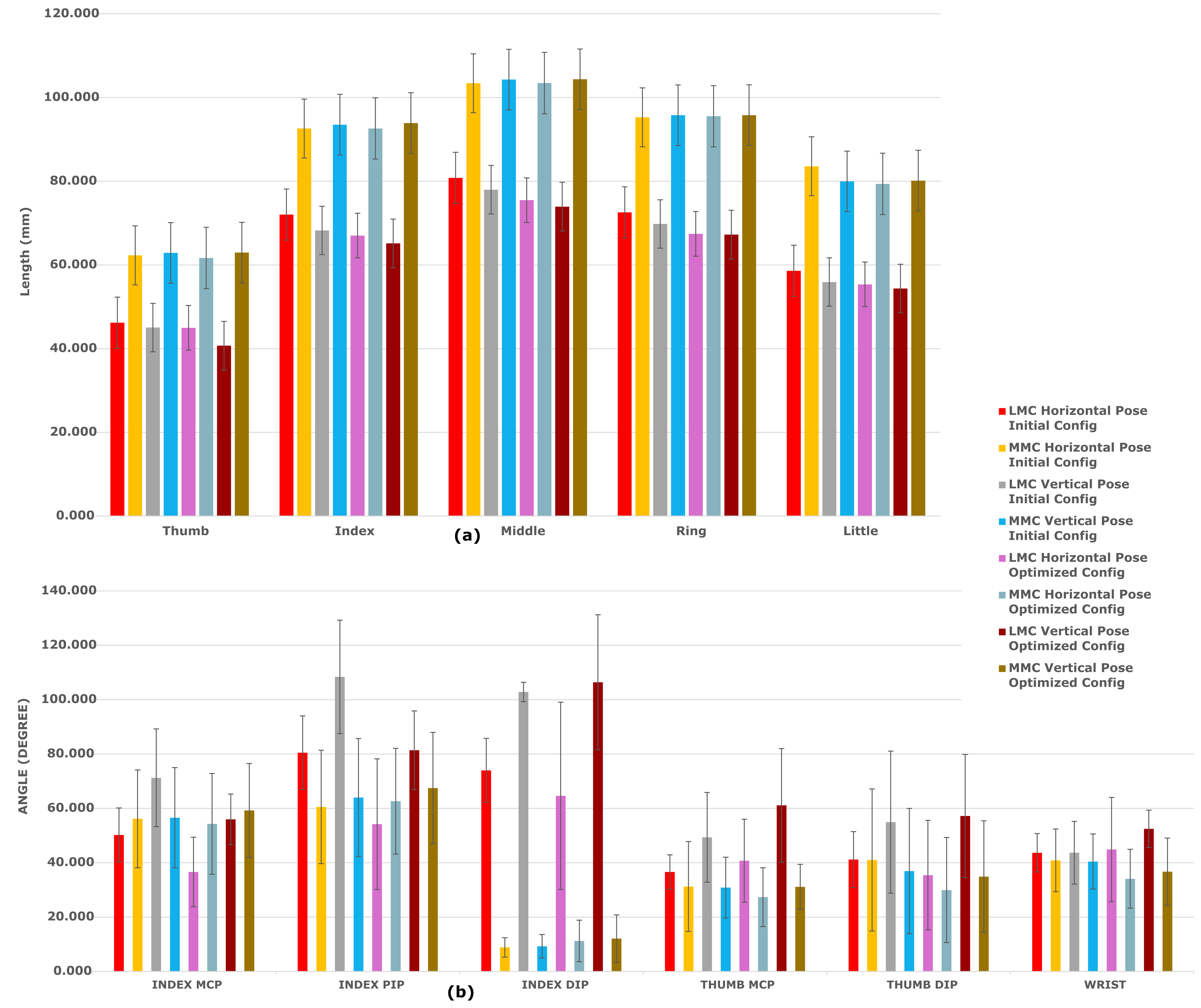}
    \vspace{-0.6\baselineskip}
    \caption{(a) Average inter-subject finger length for all static trial of each \ac{LMC} and \ac{MMC} in two configurations and hand poses. (b) Average inter-subject joint angle for all dynamic trials of each \ac{LMC} and \ac{MMC} in two configurations and hand poses.}
    \label{fig: Finger_Length_ROM}
    \vspace{-1.4\baselineskip}
\end{figure*}

\subsection{Optimized \ac{LMCs} Configuration}
The result of the first step in this study method was an optimized \ac{LMCs} configuration. Figure \ref{fig: MultiLMC Config} depicts the visualization of initial and optimized configurations. Optimized figuration has a unique position where \ac{LMC}2 was located rather far forward in the Z+ axis direction: 342,57 mm. The orientation of three \ac{LMCs}: \ac{LMC}2, \ac{LMC}3, and \ac{LMC}4, which face in negative rotation of X axis: -7.57$^{\circ}$, -8.72$^{\circ}$, and -12.06$^{\circ}$ respectively, had the purpose of seeing the hand in the vertical pose. Table \ref{tab: config_coordinates} shows the coordinates of initial and optimized configurations in detail.

\begin{table}[tb]
\centering
\caption{Center point coordinates and axis rotation of \ac{LMCs}}
\label{tab: config_coordinates}
\vspace{-0.6\baselineskip}
\scriptsize
\centering
\begin{tabular}{@{}*{6}{c}@{}}
\toprule
 & \multicolumn{3}{c}{\textbf{Coordinate}} & \multicolumn{2}{c}{\textbf{Rotation against}} \\ 
 & \multicolumn{3}{c}{\textbf{(mm)}} & \multicolumn{2}{c}{\textbf{ axis (degree)}} \\ \cmidrule(lr){2-4} \cmidrule(lr){5-6}
 & \textbf{x} & \textbf{y} & \textbf{z} & \textbf{x} & \textbf{y} \\ \midrule
\textbf{Initial Configuration} & & & & & \\
LMC1 & -60 & 0 & 60 & 0 & 0 \\
LMC2 & 60 & 0 & 60 & 0 & 0 \\
LMC3 & -60 & 0 & -60 & 0 & 0 \\
LMC4 & 60 & 0 & -60 & 0 & 0 \\ \midrule
\textbf{Optimized Configuration} & & & & & \\
LMC1 & -120.37 & 0 & 256.29 & 29.90 & 16.50 \\
LMC2 & -69.97 & 0 & 342.57 & -7.57 & -7.04 \\
LMC3 & -190.60 & 0 & 88.70 & -8.72 & -4.93 \\
LMC4 & 178.88 & 0 & 100.90 & -12.06 & 12.38 \\ \bottomrule
\end{tabular}
\vspace{-1.8\baselineskip}
\end{table}

\subsection{Validity and Repeatability}

The \ac{MMC} system was employed alongside the initial and optimized configurations of \ac{LMCs} to validate the effectiveness of using multiple \ac{LMCs} for assessing hand properties. The Vicon system, serving as the \ac{MMC}, is considered the gold standard for validating the measurement outcomes of the \ac{LMC}. In this study, measurements of finger length and the \ac{RoM} for specific parts of the hand and fingers were calculated and compared.

Figure \ref{fig: Finger_Length_ROM}a shows the average inter-subject finger length across all trials. The graph shows that the result has the same trend between initial and optimized configurations. The measurement of finger length from \ac{MMC} is always more significant than from \ac{LMCs}, with a difference to ground-truth for all fingers is 24.23$\pm$6.35 mm. The vertical hand pose also has a more substantial difference compared to the horizontal hand pose in both \ac{LMCs} configurations, where the differences to ground truth are 22.93$\pm$7.11 mm and 25.53$\pm$5.18 mm, respectively. This indicates that vertical pose had presented challenges in obtaining accurate readings from the \ac{LMC}. 
Due to the distribution and unique positioning of the \ac{LMCs}, this study hypothesized that the proposed optimized configuration would yield improved marker readings, potentially leading to more accurate finger length measurements. In fact, as illustrated in Figure \ref{fig: Finger_Length_ROM}a, the optimized configuration generally resulted in less significant finger length values than the initial configuration. In the optimized configuration, the finger length differences from the ground truth are 25.82$\pm$5.91 mm. This is more significant than the initial configuration differences, 22.64$\pm$6.39 mm.

The result of \ac{RoM} calculation is shown in Figure \ref{fig: Finger_Length_ROM}b. The mean difference of all joint angles to ground truth is 29.49$^{\circ}$$\pm$27.85. With this difference, the result of \ac{RoM} measurement from \ac{LMC} is far from agreeable. Especially the Index \ac{DIP}, the ground-truth value of Index \ac{DIP} only has a maximum value of 12.06$^{\circ}$, where the \ac{LMC} reading is 64.61$^{\circ}$. A comparison between the initial and optimized configurations shows no improvement; both exhibit significant deviations from the ground truth values, with differences of 25.63$^{\circ}\pm$22.62 and 35.84$^{\circ}\pm$33.73, respectively.

\subsection{Visibility Rate}
The ray-tracing algorithm implemented in this study ensures that all fingers are in the line of sight of \ac{LMCs}. \ac{LMC} uses its internal models to predict all the markers of the finger. When certain parts of the hand are detected, such as the palm marker, \ac{LMC} can predict all the finger markers even though the finger is occluded. This case is defined as detected from \ac{LMC}'s internal model. Detected from the internal model can lead to a misplacement of the finger joint position. Therefore, the proposed optimized configuration aims to detect the hand in true definition. The ray-tracing algorithm could prove this detection. This study uses the term visibility rate to define whether the finger is truly detected or comes from \ac{LMC}'s internal model. Visibility rate with value -1 means that all the hand parts are undetected. This could be because the hand is outside the detection range of \ac{LMC} or full oclusion of the hand. Value 0 means that finger marker data is acquired, but comes from an internal model and is not truly detected. A visibility rate with a value of 1 means that the finger is truly detected: the finger is in the line of sight of \ac{LMC}.

\begin{figure*}[tb]
    \centering
     \includegraphics[width=0.75\linewidth]{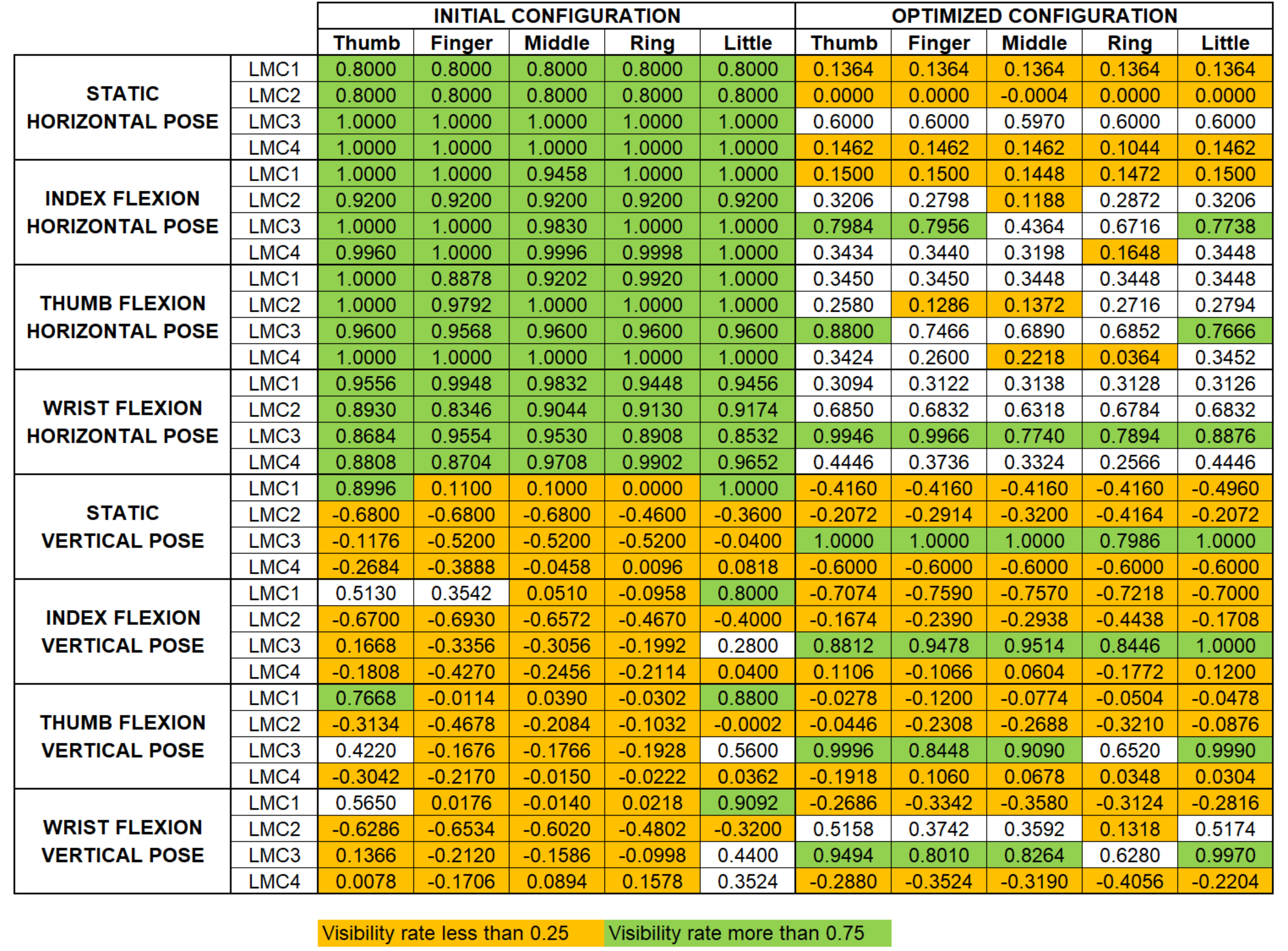}
    \vspace{-0.7\baselineskip}
    \caption{Map of mean visibility rate from ten subjects for each LMC. Visibility rate value: -1: Hand not detected, 0: Finger marker detected from \ac{LMC} internal model, 1: Finger marker truly detected from ray tracing.}
    \label{fig: Visibility Rate}
    \vspace{-1.6\baselineskip}
\end{figure*}

This study experiment has eight combinations of pose and motion that were done in initial and optimized configurations. Figure \ref{fig: Visibility Rate} shows the mean visibility rate of each \ac{LMC} across trials and subjects specific to each motion in each configuration. The result indicates that in optimized configuration, one \ac{LMC} always truly detects any movement. In the initial configuration, the value of the visibility rate is mostly less than 0.25, which means either the marker is not detected or the marker comes from the internal model of \ac{LMC}. LMC3 was the \ac{LMC} in the optimized configuration that 83.33\% can read the marker with a visibility rate of more than 0.75. It means that in optimized configuration, at least "one \ac{LMC} can truly detect" policy is fulfilled. This policy is not satisfied in the initial configuration when doing vertical pose.

\section{Discussion} 
\label{sec: Discusion}

The use of multiple \ac{LMCs} in previous research has been studied by \cite{houston2021evaluation, NovacekProjectMultileap, Kiselev2019HandGesture, WangEnlarging, Hu2017HumanStochastic}. They employed multiple \ac{LMCs} and arranged them in specific configurations to ensure accurate hand tracking. Albeit the placement is working in their study, there is no analysis or discussion regarding the optimality of the configuration. From those studies, only Houston et al. \cite{houston2021evaluation} use multiple \ac{LMCs} for measuring hand anthropometrics. Houston \cite{houston2021evaluation} placed each \ac{LMC} angled by 15$^{\circ}$ toward the hand in triangle configuration. In this study, we introduced a method to find the optimized placement for multiple \ac{LMCs} for measuring hand anthropometrics. Our result agrees with previous research carried out by Ganguly et al. \cite{ganguly2021comparison} regarding the measurement of finger length in a static pose. Although Ganguly \cite{ganguly2021comparison} uses only one \ac{LMC}, we have a similar trend where the value of finger length was always less than \ac{MMC} as the ground truth value in both initial and optimized configurations. The \ac{RoM} result significantly deviated from the ground truth, particularly for the Index \ac{DIP} joint. The \ac{LMCs} consistently recorded more significant angles for this joint, likely due to the substantial changes occurring during index flexion. The rapid movement in this joint posed challenges for our Kalman model, leading to inaccuracies in state prediction.

Employing ray-tracing for the confidence estimation in a \ac{LMC} measurement was motivated by a pilot study on a single \ac{LMC}, where the effects of occluding different hand parts were investigated. The results showed that blocking the line of sight to a single virtual marker location of a particular finger leads to immediate changes in the \ac{LMC} marker output of the same finger. The nature of the occlusion via palm, other fingers, or external objects did not influence that behaviour. Since the inner workings of the \ac{LMC} model are proprietary, the performance measure based on the visibility of virtual markers forms a lower bound on the confidence threshold. Furthermore, a \ac{LMC} might not register a hand at the borders of its \ac{FoV}, even though the ray-tracing algorithm predicted good visibility. During optimization, the maximal \ac{LMC} visibility range and the \ac{FoV} angles were thus reduced from 0.6 m to 0.4 m and from 120°/150° to 100°/100°, respectively. 

This study identified several limitations associated with the \ac{LMC}. Previous research has highlighted the ongoing issue of variable sampling with \ac{LMC} \cite{ganguly2021comparison, WangEnlarging, houston2021evaluation}. When multiple \ac{LMCs} are employed, the variable sampling rate decreases to the maximum frame rate divided by the number of \ac{LMCs}. For instance, with four \ac{LMCs} utilized, the variable sampling rate for each \ac{LMC} is capped at 30 Hz (calculated as 120 Hz divided by 4). Experimental results confirmed this limitation, with \ac{LMC} sampling rates ranging from 11 to 34 Hz. 

Another constraint observed with \ac{LMC} is axis-cutting. Although the \ac{LMC} specifications indicate an interaction zone of approximately 60 cm within a field of view (FoV) of 120$^{\circ}$x150$^{\circ}$, markers beyond 250 mm from the \ac{LMC} centre are affected. This axis-cutting phenomenon was demonstrated in the initial phase of our study. It aimed to determine the edge of the \ac{LMC} capture volume. During this experiment, the hand was moved along the x, y, and z-axes to identify areas where it was not detected. While the hand remained normal-detected within the 60 cm interaction zone, it exhibited abnormal shapes in the visualization when positioned around the 250 mm mark along any axis, despite all markers being detected. Through data inspection, the coordinate value of this axis was cut when the hand distance was more than 250 mm. This will introduce an offset if \ac{LMCs} configuration places one of the \ac{LMC} away than 250 mm. This offset needs to be included in the realignment process to solve this. On the other hand, if the hand moves further than 350 mm, all markers are still detected, but the unusual shape of the hand becomes more prominent. An error in marker reading may occur if the subject places their hands too far away. To prevent any further error introduced by this limitation, this study suggests the placement of multiple \ac{LMCs} should be around a maximum of 250 mm at any axis. The \ac{LMC} internal model of marker prediction for elbow marker was mostly incorrect to actual ground truth. Although all the hand markers were detected, the elbow marker data is not correctly predicted by the internal model of \ac{LMC}. This leads to differences in elbow marker position coming from each \ac{LMC}. In this study, the reading of the elbow marker from LMC1 has a difference of 10 - 20 mm in the y-axis after the realignment process compared to other \ac{LMCs}. This difference leads to unstable state prediction for the elbow marker in the Kalman filter process. Although \ac{LMC} has limitations, the marker read from \ac{LMC} still can be used well for hand gesture recognition. Using multiple \ac{LMC} in the optimized configuration can ensure that all the markers have better visibility than in the initial configuration.

From the optimization result of \ac{LMCs} configuration, three \ac{LMCs} are located on the left side of origin (x-) and one \ac{LMC} on the right side (x+). This placement is logical because the reference hand pose used for optimization involves the motion of the index and thumb fingers in a vertical pose that can be seen properly from the left side. To satisfy the correct marker reading, the optimized configuration gave results that LMC1, LMC2, and LMC3 were placed on the left side of the world origin coordinates. The LMC1 also has rotation in the x-axis, which ensures the fingers are fully detected when making wrist motion. In conclusion, the ray tracing algorithm will place all \ac{LMCs} in multiple configurations in a way that all fingers are in the line of sight of at least one \ac{LMC}.

For future work, the optimization component will incorporate a refined \ac{LMC} model along with additional metric terms aimed at aligning \ac{LMC} orientations to configure where trial markers are centred in the \ac{FoV}. To address bandwidth limitations and variable sampling drops in a single computation for multiple \ac{LMCs}, several strategies could be considered for future research: (1) Implementation of a multi-threaded process for computation and online Kalman filtering. Currently, the computation in this study's program occurs within the same thread as data reading from \ac{LMCs}. Separating these tasks into different threads can reduce the load on marker reading. (2) Utilization of a computer equipped with more USB ports. Assigning each \ac{LMC} its own USB port ensures that bandwidth requirements are adequately met. (3) Exploration of a Kalman Filter model tailored specifically for \ac{LMCs}, where the relationships between marker positions are precisely defined. This Kalman Filter model should also be capable of handling missing or \ac{NaN} value to ensure that it does not adversely affect the prediction state.
\section{Conclusion} 
\label{sec: Conclusion}

This study introduces a comprehensive approach to markerless motion capture using an online multi-\ac{LMC} framework with a single computation system. It incorporates a placement optimization algorithm to determine ideal device placement based on expected \ac{LMC} frame samples during trials. The methodology is validated against the gold standard \ac{MMC}, demonstrating its effectiveness and reliability. By addressing the limitations of singular \ac{LMC} devices, our approach provides a robust solution for markerless motion capture, offering clinicians a valuable tool for assessing grasping movements and informing rehabilitation strategies effectively. In terms of ratios, the finger length measurements obtained from the initial and optimized configurations are consistent with those of the \ac{MMC} system. The ratios between fingers in both configurations match those of the \ac{MMC} across all poses. But value-wise, there is no significant improvement from the initial and the optimized configuration to the validation method using \ac{MMC} system. Both configurations also could not accurately measure \ac{RoM} from dynamics movement and have a significant deviation from the ground truth value. Although neither of the multiple \ac{LMCs} configurations fully matched the performance of the standard \ac{MMC} system, the proposed optimized \ac{LMCs} configuration demonstrated better detection of fingers during vertical pose motion. It achieved a mean visibility rate of 0.05 $\pm$ 0.55, compared to -0.07 $\pm$ 0.40 for the initial configuration. This improved detection suggests the potential for further optimization of configurations tailored to specific hand poses.


\section*{ACKNOWLEDGMENT}
This work was supported by the Federal Ministry of Education and Research of the Federal Republic of Germany (BMBF) by funding the project AI.D under Project Number 16ME0539K. The first author completed this work while receiving a scholarship funded by Indonesia Endowment Fund for Education (LPDP).
\typeout{}
\bibliographystyle{ieeetr}
\bibliography{Bibliography}

\end{document}